\documentclass{article}

\usepackage[preprint]{neurips_2023}

\usepackage[utf8]{inputenc} 
\usepackage[T1]{fontenc}    
\usepackage{hyperref}       
\usepackage{url}            
\usepackage{booktabs}       
\usepackage{amsfonts}       
\usepackage{nicefrac}       
\usepackage{microtype}      
\usepackage{xcolor}         
\usepackage{subfiles}
\usepackage{amsthm}
\usepackage{amsmath}
\usepackage{amssymb}
\usepackage{mathtools}
\usepackage{booktabs}
\usepackage{multirow}
\usepackage{subfigure}
\usepackage{csquotes}
\usepackage{algorithmic}
\usepackage{algorithm}
\theoremstyle{plain}
\newtheorem{theorem}{Theorem}[section]
\newtheorem{prop}[theorem]{Proposition}

\theoremstyle{definition}
\newtheorem{definition}[theorem]{Definition}

\theoremstyle{remark}

\newcommand{\probP}{\text{I\kern-0.15em P}}
\newcommand{\E}{\mathbb{E}}

\newcommand{\Xx}{\textbf{$\mathcal{X}$}}

\newcommand{\W}{\mathcal{W}}

\newcommand{\D}{\textbf{\textit{D}}}
\newcommand{\M}{\mathcal{M}}

\usepackage{amsmath}

\DeclareMathOperator*{\argmin}{arg\,min}
\title{Privacy Preserving Bayesian Federated Learning in Heterogeneous Settings}

\author{%
  Disha Makhija \\
  Electrical and Computer Engineering\\
  The University of Texas at Austin\\
  Austin,, TX USA \\
  \texttt{disham@utexas.edu} \\
   \And
   Joydeep Ghosh \\
   Electrical and Computer Engineering \\
   The University of Texas at Austin\\
   Austin, TX, USA \\
   \And 
   Nhat Ho \\
   Department of Statistics and Data Sciences \\
   The University of Texas at Austin\\
   Austin, TX, USA \\
}

\begin{document}

\maketitle

\begin{abstract}
In several practical applications of federated learning (FL),  the clients are highly heterogeneous in terms of both their data and compute resources, and therefore enforcing the same model architecture for each client is very limiting. Moreover, the need for uncertainty quantification and data privacy constraints are often particularly amplified for clients that have limited local data. This paper presents a unified FL framework to simultaneously address all these constraints and concerns, based on training customized local Bayesian models that learn well even in the absence of large local datasets. A Bayesian framework provides a natural way of incorporating supervision in the form of prior distributions. We use priors in the functional (output) space of the networks to facilitate collaboration across heterogeneous clients. Moreover, formal differential privacy guarantees are provided for this framework. Experiments on standard FL datasets demonstrate that our approach outperforms strong baselines in both homogeneous and heterogeneous settings and under strict privacy constraints, while also providing characterizations of model uncertainties.


\end{abstract}

\section{Introduction}

The vast majority of research on Federated Learning (FL) takes an optimization problem perspective i.e., focus on designing suitable optimization objectives or provide methods for efficiently solving such objectives. However there are several critical applications where other issues such as obtaining good estimates of the uncertainty in an outcome or data privacy, are equally pressing. For example, in certain healthcare applications, patient data is private and uncertainty around prediction of health outcomes is required to manage risks. Similarly, for several applications in legal, finance, mission critical IoT systems~\citep{iout}, etc., both the privacy and confidence guarantee are important for decision making. 

Since Bayesian learning methods are known for their ability to generalize under limited data, and can generally provide well-calibrated outputs~\citep{NIPS2016_bcc0d400, gp_tree, pFedGP}, it makes them suitable for solving the challenges mentioned above. A straightforward Bayesian FL method would perform the following steps - each client does local posterior inference to obtain a distribution over weight parameters and then communicates the local posteriors to the server, the server receives the local posteriors from the clients and aggregates them to obtain a global posterior distribution which is then broadcast to the clients for the next round of training. However, this entire learning procedure is highly resource and communication intensive. For a very simple example, solving an $m$-dimensional federated least squares estimation, this method will require $O(m^3)$ computation on all the clients and server sites~\citep{fedPA}. This cost is much more as opposed to the standard FL which is generally $O(m)$. How could the Bayesian methods be then used for FL settings without paying such high costs? Also, since clients can have varied degrees of compute resources, constraining each client to train identical models would be limiting. Then, how to aggregate the local posteriors when the clients are training personal models and the model weights across clients are not identical, becomes the next important question. We try to address these questions by proposing a framework that allows all clients to train their own personal Bayesian models (with varying model complexities), but still enables collaboration across clients by instilling information from the peer collaboration in the form of priors. This is achieved by using  the functional space to indirectly determine priors on weights, as opposed to them being specified in the weight space as in traditional Bayesian methods. 

In addition, maintaining client data privacy is an important concern in FL. While attacks on neural networks that can recover training data from the models by either directly analyzing model parameters or indirectly by analyzing the model outputs are well known~\citep{model_attacks, pate}, even in federated learning settings, when the data is restricted to the local client sites, it has been shown that not all FL methods are immune to data leaks~\citep{robbing_the_fed, fishing_user_data, fowl2023decepticons}. Therefore, to guarantee the privacy of the local client data, we apply a formal well-known standard of differential privacy~\citep{dp}. We use a carefully designed noise mechanism for data sharing that allows us to provide a privacy bound on the entire procedure.

In this work, we propose a novel unified FL framework that can simultaneously address challenges arising due to limited data per client and heterogeneous compute resources across clients, together with the need for privacy guarantees and providing calibrated predictions. To the best of our knowledge, no previous work has jointly addressed all these learning issues in the FL context. Our positive results substantially increase the scope of FL for critical real-world applications.

\textbf{Our key contributions} are summarized as follows :
\begin{enumerate}
    \item We propose a novel computation and communication efficient method for federated learning based on Bayesian inference. In this method, the collaboration across clients is achieved through a novel way of assigning prior distributions over the model parameters via the output space. We call this method Federated Bayesian Neural Networks (FedBNN).
    \item We provide a differential privacy guarantee for our method and show that the method is able to learn effectively even under strict privacy guarantees. 
    \item We then present a non-trivial extension of our method that can work with heterogeneous clients and show that even under this extension we can achieve differential privacy. 
    \item We evaluate our method on several datasets and show that it  outperforms the baselines by a significant margin, particularly in  heterogeneous settings, making it particularly useful for FL applications where a high degree of heterogeneity is naturally present.
\end{enumerate}

\section{Related Work} \label{sec:lit_review}
This section provides an overview of the most relevant prior work in the fields of federated learning, Bayesian FL, and Differential Privacy in FL.

\vspace{0.5 em}
\noindent
\textbf{Federated Learning} In the early 2000s, privacy preserving distributed data mining referred to training distributed machine learning models~\citep{kapa00, gali17, aggarwal2008general}, like distributed clustering~\citep{megh03, megh05}, distributed PCA~\citep{kahu01}, distributed SVMs~\citep{yuji06} etc. Federated Learning was introduced as the FedAvg algorithm in the seminal work by~\citep{fedavg}. Since then many different modifications have been proposed that tackle specific challenges. FedPD~\citep{fedpd}, FedSplit~\citep{fedsplit}, and FedDyn~\citep{feddyn} proposed methods for finding better fixed-point solutions to the FL optimization problem.~\citep{fedDF, pfnm, fedma, singh2020model, fedbe} show that point-wise aggregate of the local client models does not produce a good global model and propose alternate aggregation mechanisms to achieve collaboration. FedDF~\citep{fedDF} achieves collaboration by performing knowledge distillation on local client models. While the above mentioned methods work on creating a global solution, it has also been shown that personal models achieve better performance for local clients' tasks. Personalised FL has been approached in many different ways like meta-learning~\citep{personalised_meta_learning, waffle, maml, KhodakBT19}, multi-task learning~\citep{fed_mtl, ditto, mocha}, by clustering the clients~\citep{SattlerMS21, GhoshCYR20} and others~\citep{fedrep, fedprox, local_adaption, pfedme, flexifed, Makhija2022ArchitectureAF}. The personalised FL methods focus on improving the performance in the presence of statistical data heterogeneity across clients, but they do not work well when the size of dataset on the clients is limited.

\vspace{0.5 em}
\noindent
\textbf{Bayesian Federated Learning} The Bayesian approaches for federated learning can be broadly divided into two categories - methods using Bayesian inference for local learning at each client and methods that achieve collaboration via Bayesian mechanisms. Amongst the methods that use Bayesian approach for achieving collaboration across clients, FedBE~\citep{fedbe} uses Bayesian mechanism to aggregate the locally trained neural networks to obtain a Bayesian ensemble at the server.~\citep{fedPPD} suggests using knowledge distillation and MCMC based method for training a global model.~\citep{thompson_sampling} on the other hand suggests the use of Bayesian Optimization and Thompson Sampling to obtain the solution to the global optimization problem. PFNM~\citep{pfnm} uses a Beta-Bernoullli process to obtain a global model from the local models but is only applicable to fully-connected networks, FedMA~\citep{fedma} extends PFNM for other types of networks. Recently, ~\citep{ozer2022how} did an empirical study on various ways of aggregating variational Bayesian neural networks and their effects. On the other hand, FedPA~\citep{fedPA} was the first to use local Bayesian inference and suggested that the global posterior distribution of weights could be obtained by multiplying local posteriors and proposed an approximate and efficient way of computing local and global posteriors by using Laplace approximations with complexity that is linear in number of model parameters. This method focuses on obtaining a global solution and is less suited for the statistical heterogeneity present across clients~\citep{bfl_survey}, and therefore we focus more on the methods that build personalised solutions for clients. Among such methods, pFedGP~\citep{pFedGP} is a Gaussian Process based estimation method where the kernel in these GPs is defined on the output of a neural network and is called Deep Kernel Learning. pFedGP works by collaboratively training a single deep kernel for all clients but using personalised GPs for prediction. The collaboration across clients while learning the global kernel is done as in FedAvg. FedLoc~\citep{fedloc} also uses GP in FL but for regression tasks. pFedBayes~\citep{pfedbayes} uses variational inference for local posterior inference on Bayesian Neural Networks(BNNs) where the loss at each client is a combination of the data likelihood term and distance to the prior where prior distribution is replaced by the global distribution. The global distribution is obtained by aggregating the local prior distributions. FOLA~\citep{fedLA} proposed using Laplace Approximation for posterior inference at both the server side and the client side. None of the methods described above explicitly handle heterogeneous settings. Moreover, for these methods choosing an appropriate prior distribution over the local model parameters is a challenge~\citep{bfl_survey}. These issues  led us to using functional space priors instead. Such priors have been studied in limited centralised settings~\citep{Tran2022, sun2018functional, FlamShepherd2017MappingGP} but not in FL settings.

\vspace{0.5 em}
\noindent
\textbf{Differential Privacy in FL} Since decentralised learning does not guarantee that the data will remain private, it is important that a formal rigorous guarantee be given on the data that is leaked by the algorithm. Seminal works in DP propose using a Gaussian noise mechanism by adding Gaussian noise to the intermediate results and achieving a bound on the algorithm by using composition results~\citep{dp, renyi_dp, dp_composition}. For FL,~\citep{dp_fl} and~\citep{dp_fedavg} independently proposed DP-FedSGD and DP-FedAvg algorithms which enhance FedAvg by adding Gaussian noise to the local client updates. Several other works focus on analysing the privacy-utility trade-off in DP in FL setting~\citep{Ghazietal, girgis, balle_et_al, bayesian_dp, li_et_al}. Recently,~\citep{hu_et_al} proposed a DP based solution for personalised FL which only works for linear models. And then~\citep{dp_scaffold} enhanced it for general models and heterogeneous data in FL. These methods, however, mostly focus on the privacy guarantees while solving the FL optimization problem. They don't apply to  general Bayesian settings or when substantial heterogeneity is encountered.  

\section{Proposed Methodology} \label{sec:methodology}
We propose a novel method for Bayesian FL that enables privacy preserving personalised learning on clients under heterogeneous settings. The local training is done in a model agnostic way but collaboration amongst clients is enabled by passing information as the prior. And to achieve privacy, like most private algorithms, we use Gaussian mechanism to carefully add noise to the information sent outside a client. This method also produces well calibrated outputs. In this section, we first describe the problem statement and then go over the proposed solution.

\subsection{Problem Description}
Consider an FL setting with $N$ clients where each client $i$ has local dataset $\Xx_i$ of size $n_i$ drawn from the local data distribution $\D_i$. The goal of a personalised federated learning procedure is to obtain optimal weights for each client's local model, $\W_i$, using data $\Xx = \bigcup_{j=1}^{N}\Xx_j$ without actually accessing the data outside of each client, i.e., no other client can access any data in $\Xx_i$ except the $i^{th}$ client itself but the clients could transfer knowledge via collaboration. In a personalised Bayesian learning procedure, the modified goal would be to learn distribution over local weights, $\probP(\W_i |\Xx)$ from $\Xx = \bigcup_{j=1}^{N}\Xx_j$ while still maintaining the client data privacy. However, the learning procedure faces challenges that are posed due to - \textit{system heterogeneity} and \textit{statistical heterogeneity}. System heterogeneity refers to the variable amount of data and compute resources across clients, meaning, i) the data resources on each client vary widely, i.e., $n_k >> n_l$ for some clients $k$ and $l$, and ii) the compute across clients is non-identical due to which it is not possible to train models of uniform architectures across clients, leading to non-identical weights, i.e., $\mathcal{W}_i \neq \mathcal{W}_j$ for different clients $i$ and $j$. Statistical heterogeneity implies that the data distribution across clients is non-IID. 

\subsection{FedBNN Methodology}
Here we propose an adaptive framework to learn personalised Bayesian Neural Network (BNN) based models for each client that can work effectively even under the challenges described above. The overall framework works iteratively in two steps - local optimization on the individual client to obtain local posterior distribution over the model parameters, and a global collaboration step where the output from each client is appropriately aggregated at the server and broadcast to all the clients for the next rounds of training. Each of these two steps is described in detail below, and the detailed algorithm is given in the Appendix.

\vspace{0.5 em}
\noindent
\textbf{Local Setting} Let each client in the network be training a personalised Bayesian NN, which for the client $i$ is denoted by $\Phi_i$ and is parameterised by weights $\mathcal{W}_i$. As commonly used in the literature, we assume that the individual weights of the BNN are Normally distributed and satisfy mean-field decomposition, i.e., $w_{i,\alpha} \sim \mathcal{N}(\mu_{i,\alpha}, \sigma^2_{i,\alpha})$ for $\alpha \in [1, \dots, |\mathcal{W}_i|]$ where $\mu_{i,\alpha}$ is the mean of the gaussian distribution for the parameter $\alpha$ on the $i^{th}$ client and $\sigma^2_{i,\alpha}$ is the variance of the gaussian for the same parameter. To guarantee that $\sigma_{i,\alpha}$ takes non-negative values for all clients $i$ and all parameters $\alpha$, we use a technique commonly used in inference procedures~\citep{bbb}, wherein each $\sigma_{i,\alpha}$ is replaced by another parameter $\rho_{i,\alpha}$ during the training, such that $\sigma_{i,\alpha} = \text{log}(1 + \text{exp}(\rho_{i,\alpha}))$. 

\subsubsection{Global Collaboration} We attain collaboration amongst clients under a mild assumption of availability of a general publicly accessible unlabelled dataset at the server. There is no assumption on the distribution of this dataset as the clients' data distributions are anyway unknown in the real-world. We call this dataset as Alignment Dataset (AD). This dataset is used for providing peer supervision to the clients by helping clients distill knowledge from other clients without sharing the model weights or the local posterior distributions. Because in heterogeneous settings, with non-identical models across clients, working and aggregating information in the weight space becomes difficult and non-intuitive, for the purpose of collaboration, we move away from the weight-space to the function-space of the networks. Specifically, in each global communication round, the server shares the AD with all the clients. The clients do a forward pass on AD to obtain the output $\Phi_i(\text{AD})$. The local output of the $i^{th}$ client is approximated by doing Monte Carlo sampling and drawing $K$ weight samples, $\W_i^{(j)} : j \in [1,K]$, from its local posterior distribution $\probP(\W_i |\Xx)$. An aggregate of the obtained logits for these $K$ weight samples under the client's own personal BNN model, $\Phi_i()$, is reported, i.e. $\Phi_i(\text{AD}) = \dfrac{1}{K}\sum_{j=1}^{K} \Phi_i(\text{AD}; \W_i^{(j)})$. The obtained output for AD on each client is then sent back to server which forms an aggregated representation, denoted by $\Bar{\Phi}(\text{AD})$, by doing a weighted aggregation of all clients' outputs, i.e., $\Bar{\Phi}(\textbf{X}) =  \sum_{j=1}^{N} w_j \Phi_{j}(\textbf{X}).$ The weights $w$'s used in the aggregation could represent the strength of that particular client in terms of its data or compute resources, i.e., clients with high compute (or data) resources receive more weight as compared to clients with lower amount of resources. The obtained $\Bar{\Phi}(\text{AD})$ is then uploaded to all the clients for use in the next round of local training. More details about the Alignment Dataset (AD) are included in the Appendix.

\subsubsection{Local Optimization on Clients}
\paragraph{Prior Specification} The Bayesian framework provides a natural way of incorporating supervision in the form of priors. Conventional methods in Bayesian deep learning provide priors over model parameters. However, the relationship between the model parameters and the outputs is complex and the priors in model's weight-space do not directly capture the desired functional properties. Also, since the number of parameters in a neural network is large, most prior specifications tend to take a simplistic form like an isotropic Gaussian, to make inference feasible. Thus, learning by specifying prior distributions over weights does not always help incorporate prior knowledge in the learning process. In this work, we consider a way of specifying priors in the functional space by first optimising the Bayesian neural networks over the prior parameters for a fixed number of steps to achieve a desired functional output. While being more intuitive, these priors also help in instilling the prior external knowledge during the training of the neural networks.

\vspace{0.5 em}
\noindent
\textbf{Local Optimization} For the local optimization, the individual clients learn $\probP(\mathcal{W}_i | \Xx_i)$ via variational inference. A variational learning algorithm, tries to find optimal parameters $\theta^*$ of a parameterized distribution $q(\mathcal{W}_i | \theta)$ among a family of distributions denoted by $\mathcal{Q}$ such that the distance between $q(\mathcal{W}_i | \theta^*)$ and the true posterior $\probP(\mathcal{W}_i | \Xx_i)$ is minimized. In our setting, we set the family of distributions, $\mathcal{Q}$, to be containing distributions of the form $w_{i,\alpha} \sim \mathcal{N}(\mu_{i,\alpha}, \sigma^2_{i,\alpha})$ for each parameter $w_{i,\alpha}$ for $\alpha \in [1, \dots, |\mathcal{W}_i|]$. Let $p(\W_i; \psi)$ represent the prior function over the weights $\W_i$ and is parameterized by $\psi$, then the optimization problem for local variational inference is given by :
\begin{align}\label{eqn:prior_params}
    \theta^* &= \argmin_{\theta : q(\mathcal{W}_i | \theta) \in \mathcal{Q}} \text{KL}[q(\mathcal{W}_i | \theta) || \probP(\W_i | \Xx_i)] \\
    &= \argmin_{\theta : q(\mathcal{W}_i | \theta) \in \mathcal{Q}} \text{KL}[q(\W_i | \theta) || p(\W_i; \psi)] - \E_{q(\mathcal{W}_i | \theta)}[\text{log} \probP(\Xx_i | \W_i)].
\end{align}

For inference in Bayesian neural networks, we use Bayes by Backprop~\citep{bbb} method to solve the variational inference optimization problem formulated above. 

Before beginning the local optimization procedure, we use the global information obtained from the server $\Bar{\Phi}(\text{AD})$ to intialize the prior for the BNN. Specifically, at the beginning of each local training round, the selected clients first tune their priors to minimize the distance between the local output, $\Phi_{i}(\textbf{\text{AD}}; \W_i)$ and the aggregated output obtained from the server, $\Bar{\Phi}(\text{AD})$. Since the aggregated output represents the collective knowledge of all the clients and may not be \textit{strictly precise} for the local model optimization, we consider this aggregated output as ``noisy" and correct it before using for optimization. Specifically, we generate $\Phi_{i}^{\text{corrected}}$ as a convex combination of the global output and the local output for a tunable parameter $\gamma$. For the $i^{th}$ client,
\begin{align}
\Phi_{i}^{\text{corrected}} = \gamma \Bar{\Phi}(\text{AD}) + (1 - \gamma) \Phi_{i}(\text{AD}; \mathcal{W}_i).
\end{align}
The prior optimization steps then optimize the distance between $\Phi_{i}^{\text{corrected}}$ and $\Phi_{i}(\text{AD}; \mathcal{W}_i)$ to train the prior parameters $\psi$, with the aim of transferring the global knowledge encoded in $\Phi_{i}^{\text{corrected}}$ to the local model. Precisely,
\begin{align}\label{eqn:psi}
\psi^* = \argmin_{\psi } \text{d}(\Phi_{i}^{\text{corrected}}, \Phi_{i}(\text{AD}; \mathcal{W}_i)).
\end{align}
When the outputs $\Phi(\text{X}; \W)$ are logits, we use cross-entropy or the negative log-likelihood loss as the distance measure. The optimization involves training the client's personal BNN $\Phi_i$ to only learn the parameters of the prior distribution denoted by $\psi$. This way of initializing the BNN prior enables translating the functional properties, as captured by $\Phi_{i}(\text{AD}; \mathcal{W}_i))$, to weight-space distributions. The optimal prior parameters are then kept fixed while training the BNN over the local dataset. The local optimization procedure now works to find the best $ q(\mathcal{W}_i | \theta)$ fixing the prior distribution through the following optimization problem :
\begin{align}
\theta^* = \argmin_{\theta : q(\mathcal{W}_i | \theta) \in \mathcal{Q}} \text{KL}[q(\W_i | \theta) || p(\W_i; \psi^*)] - \E_{q(\mathcal{W}_i | \theta)}[log \probP(\Xx_i | \W_i)].
\end{align}

\subsubsection{Achieving Differential Privacy} In this method, we measure the privacy loss at each client. To control the release of information from the clients, we add a carefully designed Gaussian mechanism wherein we add Gaussian noise to the $\Phi_i(\text{AD})$ that is being shared by each client. Specifically, each client $i$ uploads $\Phi_i(\text{AD})_{\text{DP}} = \Phi_i(\text{AD}) + \mathcal{N}(0, \sigma_g^2)$ to the server and then the server aggregates $\Phi_i(\text{AD})_{\text{DP}}$ across clients to obtain and broadcast $\Bar{\Phi}(\text{AD})_{\text{DP}}$ which is used by the clients in their next round of local optimization. The variance of the noise depends on the required privacy guarantee. 

\section{Privacy Analysis} \label{sec:privacy_analysis}
Since our algorithm does not specifically share model weights outside of the clients but shares the client models' output on a public dataset, it might seem intuitive that the algorithm is private but knowing that privacy is lost in unexpected ways, we present a formal Differential Privacy based guarantee on the algorithm. Our analysis in this section focuses on providing record-level DP guarantee over the entire dataset $\Xx$. This analysis quantifies the level of privacy achieved towards any third party and an honest-but-curious server. In this section we first go over some DP preliminaries and then present our analysis.

\begin{definition}[$(\epsilon, \delta)$- Differential Privacy]\label{def:dp}
A randomized algorithm $\M$ is $(\epsilon, \delta)$-DP if for any two neighboring datasets $D$ and $D'$ that differ in at most one data point, the output of the algorithm $\M$ on $D$ and $D'$ is bounded as
$$
\probP[\M(D) \in S] \leq \text{e}^{\epsilon}\probP[\M(D') \in S] + \delta, \quad \forall S \subseteq \text{Range}(\M).
$$
\end{definition}
A generalization of differential privacy is concentrated differential privacy(CDP), an alternative form of CDP called zero-concentrated differential privacy(zCDP) was proposed to enable tighter privacy analysis~\citep{dp_bun}. We will also use the zCDP notion of privacy for our analysis. The relationship between standard DP and zCDP is shown below.

\begin{prop}[$(\epsilon, \delta)$-DP and $\rho$-zCDP]\label{prop:zcdp_dp}
For a randomized algorithm $\M$ to satisfy $(\epsilon, \delta)$-DP, it is sufficient for it to satisfy $\frac{\epsilon^2}{4 \text{log}\frac{1}{\delta}}$-zCDP. And a randomized algorithm $\M$ that satisfies $\rho$-zCDP, also satisfies $(\epsilon', \delta)$-DP where $\epsilon' = \rho + \sqrt{4 \rho \text{log}\frac{1}{\delta}}$.
\end{prop}
As opposed to the notion of DP, the zCDP definition provides tighter bounds for the total privacy loss under compositions, allowing better choice of the noise parameters. The privacy loss under the serial composition and parallel composition incurred under the definition of zCDP was proved by~\citep{dp2} and is recalled below.

\begin{prop}[Sequential Composition]\label{prop:sequential_composition}
Consider two randomized mechanisms, $\M_1$ and $\M_2$, if $\M_1$ is $\rho_1$-zCDP and $\M_2$ is $\rho_2$-zCDP, then their sequesntial composition given by $(\M_1(), \M_2())$ is $(\rho_1 + \rho_2)$-zCDP.
\end{prop}

\begin{prop}[Parallel Composition]\label{prop:parallel_composition}
Let a mechanism $\M$ consists of a sequence of $k$ adaptive mechanisms, $(\M_1, \M_2, \dots \M_k)$ working on a randomized partition of the $D = (D_1, D_2, \dots D_k)$, such that each mechanism $\M_i$ is $\rho_i$-zCDP and $\M_t : \prod_{j=1}^{t-1}\mathcal{O}_j \times D_t \rightarrow O_t$, then $\M(D) = (\M_1(D_1), \M_2(D_1), \dots \M_k(D_k))$ is $\max_i \rho_i$-zCDP.
\end{prop}

After computing the total privacy loss by an algorithm using the tools described above, we determine the variance of the noise parameter $\sigma$ the total loss with a set privacy budget. The relationship of the noise variance to privacy has been shown in prior works by~\citep{dp, dp2} and is given below.
\begin{definition}[$L_2$ Sensitivity]\label{def:l2_sens}
    For any two neighboring datasets, $D$ and $D'$ that differ in at most one data point, $L_2$ sensitivity of a mechanism $\M$ is given by maximum change in the $L_2$ norm of the output of $\M$ on these two neighboring datasets
    $$ \Delta_2(\M) = \sup_{D, D'} || \M(D) - \M(D') ||_2. $$
\end{definition}

\begin{prop}[Gaussian Mechanism]\label{prop:gaussian_mechanism}
    Consider a mechanism $\M$ with $L_2$ sensitivity $\Delta$, if on a query $q$, the output of $\M$ is given as $\M(x) = q(x) + \mathcal{N}(0, \sigma^2)$, then $\M$ is $\frac{\Delta^2}{2\sigma^2}$-zCDP.
\end{prop}
Equipped with the above results, we now state the privacy budget of our algorithm.
\begin{theorem}[Privacy Budget]
The given algorithm is $(\epsilon, \delta)$-differentially private, if the total privacy budget per global communication round per query is set to
$$
\rho = \dfrac{\epsilon^2}{4 E K \text{log}\frac{1}{\delta}}
$$
for $E$ number of global communication rounds and $K$ number of queries to the algorithm per round.
\end{theorem}
\begin{proof} 
    After using Gaussian mechanism on each client and adding noise to each coordinate of $\Phi_i(\text{AD})$, the local mechanism at each client becomes $\rho$-zCDP for $\rho = \frac{\Delta^2}{2\sigma^2}$. Since each client outputs the logit representation for each input, $\Delta^2 \leq 2$. Suppose in each global communication round we make $K$ queries to each client, then by sequential composition~\ref{prop:sequential_composition}, we get $E K\rho$, for $E$ number of global communication rounds. By parallel composition~\ref{prop:parallel_composition}, the total privacy loss for all $N$ clients is the maximum of the loss on each client and therefore  
    remains $E K\rho$. Relating it to the $(\epsilon, \delta)$-DP~\ref{prop:zcdp_dp}, we get $\rho = \dfrac{\epsilon^2}{4 E K \text{log}\frac{1}{\delta}}$ for any $\delta > 0$.
\end{proof}
Our analysis does not assume any specifics of how each client is trained and is therefore applicable in more general settings. However, we present a pessimistic analysis by assuming that a change in single data point may entirely change the output of the algorithm and bound the $\Delta^2 \leq 2$. Also, since the public dataset remains common throughout the rounds, the loss due to querying on the public dataset does not add up linearly. Yet the above analysis shows that we have several knobs to control to achieve the desired privacy-utility trade off - balancing the number of global communication rounds with local epochs, reducing the number of queries, and the standard noise scale. By appropriately tuning these controls we are able to achieve good performance with a single digit $\epsilon$ privacy guarantee.

\section{Experiments} \label{sec:experiments}
In this section, we present an experimental evaluation of our method and compare it with different baselines under diverse homogeneous and heterogeneous client settings. We will also discuss the change in performance of our method when the degree and type of heterogeneity, including heterogeneity in compute resources, data availability and data distribution statistics, changes. Due to the space constraint, some additional experimental results and ablation studies are included in the Appendix.

\subsection{Experimental Details}
\paragraph{Datasets} We choose three different datasets commonly used in prior federated learning works from the popular FL benchmark, LEAF~\citep{caldas2019leaf} including MNIST, CIFAR-10 and CIFAR-100. MNIST contains 10 different classes corresponding to the 10 digits with 50,000 28$\times$28 black and white train images and 10,000 images for validation. CIFAR-10 and CIFAR-100 contain 50,000 train and 10,000 test colored images for 10 classes and 100 classes respectively. 

\vspace{0.5 em}
\noindent
\textbf{Simulation Details} We simulate three different types of heterogeneous settings - corresponding to heterogeneity in compute resources, data resources and the statistical data distribution. Before starting the training process, we create $N$ different clients with different compute resources by randomly selecting a fraction of clients that represent clients with smaller compute. Since these clients do not have large memory and compute capacity, we assume that these clients train smaller size BNNs as opposed to the other high-capacity clients that train larger VGG based models. In particular, the small BNNs were constructed to have either 2 or 3 convolution layers, each followed by a ReLU and 2 fully-connected layers at the end, and a VGG9 based architecture was used for larger BNNs. The number of parameters in smaller networks is around 50K and that in larger networks is around 3M. Since the baselines only operate with identical model architectures across clients, we use the larger VGG9 based models on the baselines for a fair comparison. We include the results of our method in both homogeneous compute setting (similar to baselines) as well as in heterogeneous compute setting wherein we assume that 30\% of the total clients have smaller compute and are training smaller sized models.  

Next, we also vary the data resources across clients and test the methods under 3 different data settings - small, medium and full. The small setting corresponds to each client having only 50 training data instances per class, for the medium and full settings each client has 100 data instances and 2500 data instances per class respectively for training. We simulate statistical heterogeneity by creating non-IID data partitions across clients. We work in a rather strict non-IID setting by assuming clients have access to data of disjoint classes. For each client a fraction of instance classes is sampled and then instances corresponding to the selected classes are divided amongst the specific clients. For the included experiments, we set number of clients $N = 20$ and divide the instances on clients such that each client has access to only 5 of the 10 classes for MNIST and CIFAR-10, and 20 out of 100 classes for CIFAR-100.  

\begin{table}[!t]
\caption{Test accuracy comparsion with baselines in non-IID settings.}
\label{table:acc_noniid}
\setlength\tabcolsep{3pt}
\begin{center}
\begin{small}
\scalebox{0.9}{
\hspace*{-1.5cm}
\begin{tabular}{lccccccccc}
\toprule
 \multirow{2}{*}[-0.5ex]{Method} & \multicolumn{3}{c}{MNIST} & \multicolumn{3}{c}{CIFAR10} & 
 \multicolumn{3}{c}{CIFAR100} \\
   \cmidrule(lr){2-4} \cmidrule(lr){5-7} \cmidrule(lr){8-10}
        & (small) & (medium) & (full) & (small) & (medium) & (full) & (small) & (medium) & (full) \\[0.5ex]
 \midrule
 (Non-Bayesian) & & & & & & & & & \\
 FedAvg &  88.2 $\pm$ 0.5  &  90.15 $\pm$ 1.2   &  92.23 $\pm$ 1.1  &  43.14  $\pm$ 1.2  &  56.27  $\pm$ 1.8  &  78.17  $\pm$ 1.2  & 27.3   $\pm$ 1.9  &  32.81  $\pm$ 1.6  &  36.3  $\pm$ 0.2 \\ [0.3ex]
 FedProx &  86.9 $\pm$ 0.8  & 89.91 $\pm$ 0.7  &  93.1  $\pm$ 0.4  &  44.27  $\pm$ 1.2  & 58.93   $\pm$ 0.9  &  79.19  $\pm$ 0.6  &    28.6 $\pm$ 2.7  &  34.31  $\pm$ 1.4  &  37.8 $\pm$ 0.9 \\ [0.3ex]
 pFedME &  91.95 $\pm$ 2.1  & 93.39 $\pm$ 1.2  &  95.62  $\pm$ 0.5  & 48.46 $\pm$ 1.5  &  64.57 $\pm$ 2.1  &  75.11 $\pm$ 1.2  &  32.4  $\pm$ 2.2  &  36.3  $\pm$ 2.0  &  41.8  $\pm$ 1.7 \\ [0.3ex]
 \midrule
 \multicolumn{3}{l}{(Bayesian with Homogeneous Architectures)} & & & & & & & \\[0.3ex]
 pFedGP &  86.15 $\pm$ 1.3   & 90.59 $\pm$ 1.7  &  94.92  $\pm$ 0.3  &  45.62  $\pm$ 2.2  &  56.24  $\pm$ 1.8  &  72.89  $\pm$ 0.7  &  47.06  $\pm$ 1.3  &   53.1 $\pm$ 1.2  &  54.54  $\pm$ 0.2 \\ [0.3ex]
 pFedBayes & 94.0 $\pm$ 0.2   & 94.6 $\pm$ 0.1  &  95.5  $\pm$ 0.3  &  58.7  $\pm$ 1.1  &  64.6  $\pm$  0.8 &  78.3  $\pm$ 0.5  &   39.51 $\pm$ 1.8  &  41.43  $\pm$ 0.4  &  47.67  $\pm$ 1.1\\ [0.3ex]
 FOLA & 91.74 $\pm$ 1.0   & 92.87 $\pm$ 0.8  &  95.12  $\pm$ 0.6  &  43.29 $\pm$ 0.9  &  45.94  $\pm$ 0.7  & 67.98  $\pm$ 0.5  &  33.42  $\pm$ 1.3  &  48.8  $\pm$ 2.1  &  43.2  $\pm$ 1.6 \\ [0.3ex]
 \midrule
 Ours (Homo) & \textbf{94.9 $\pm$ 1.0}  &  \textbf{95.72 $\pm$ 0.8}   &   \textbf{96.21 $\pm$ 0.3}   &    \textbf{70.6 $\pm$ 1.1}   &  \textbf{72.3 $\pm$ 0.6}  &  \textbf{79.7 $\pm$ 0.3}   &  \textbf{49.65  $\pm$ 1.4}  &   \textbf{55.4 $\pm$ 0.8}  &  \textbf{57.3 $\pm$ 0.8}\\ [0.3ex]
 Ours (Hetero) & 93.1 $\pm$ 1.1   & 94.4 $\pm$ 0.2  &  95.9  $\pm$ 0.2  &  \textbf{68.17 $\pm$ 2.0}   &   \textbf{71.73 $\pm$ 1.3}  &  \textbf{78.7 $\pm$ 0.7}   & \textbf{47.5 $\pm$ 1.4}  &  49.10 $\pm$ 1.1  & \textbf{51.1 $\pm$ 0.7} \\ [0.3ex]
 Ours (Hetero-DP) & 89.82 $\pm$ 2.3   &  90.21 $\pm$ 1.6   &  91.43  $\pm$ 1.4  &  60.4  $\pm$ 1.1  &  68.13  $\pm$ 1.2  &  74.3  $\pm$ 1.6  &  43.7 $\pm$ 2.3  &  44.5  $\pm$  1.7 &   47.0 $\pm$ 1.5 \\ [0.3ex]
 \midrule
 (DP-Baseline) & & & & & & & & & \\
 DP-FedAvg &  80.1 $\pm$ 1.7   & 85.2 $\pm$ 1.8 &  86.2  $\pm$ 1.7  &  35.17  $\pm$ 0.8  &  50.22  $\pm$ 1.1  &  74.6  $\pm$ 1.2  &  26.5  $\pm$ 0.3  &  30.7  $\pm$ 1.4  &  32.4  $\pm$ 0.6 \\ [0.3ex]
\end{tabular}
}
\end{small}
\end{center}    
\end{table}

\vspace{0.5 em}
\noindent
\textbf{Training parameters and Evaluation} We run all the algorithms for 100 global communication rounds and report the accuracy on the test dataset at the end of the $100^{th}$ round. The number of local epochs is set to 20 and the size of AD is kept as 2000. Each client is allowed to train its personal model for a fixed number of epochs, which is kept to $50$ in experiments, before entering the collaboration phase. The hyper-parameters of the training procedure are tuned on a set aside validation set. At the beginning of each global communication round, when we optimize the prior at each client according to equation~\ref{eqn:psi}, we use an Adam optimizer with learning rate=0.0001 and run the prior optimization procedure for 100 steps. Then with the optimized prior we train the local BNN using Bayes-by-Backprop, with Adam optimizer, learning rate=0.001 and batch size = 128. The noise effect $\gamma$ is selected after fine-tuning and kept to be $0.7$. All the models are trained on a 4 GPU machine with GeForce RTX 3090 GPUs and 24GB per GPU memory. For evaluation, we report the classification accuracy obtained by running the trained models on test datasets from the MNIST, CIFAR10 and CIFAR100 datasets.

\vspace{0.5 em}
\noindent
\textbf{Baselines} We compare our method against the standard non-Bayesian FL algorithms and Bayesian-FL methods that build personalised models for clients. We also show results of differentially private FedAvg algorithm under similar privacy guarantee to provide perspective on the privacy. The FL baselines include - i) FedAvg, the de-facto FL learning algorithm which trains a global model, ii) FedProx, an enhancement of the FedAvg algorithm in the presence of statistical heterogeneity across clients giving a global model, iii) pFedME, which uses personalised models on each client using Monreau envelopes in loss. The bayesian baselines include - i) pFedGP, a Gaussian process based approach that trains common deep kernels across clients and personal tree-based GPs for classification, ii) pFedBayes, uses a variational inference based approach for personalised FL by training personal models which are close to the aggregated global models, iii) FOLA, bayesian method using Gaussian product for model aggregation. And lastly, the DP baseline includes - i) DP-FedAvg, the FedAvg algorithm with gradient clipping and noise addition to the gradient at each client. The baseline experiments were performed with the best parameters reported in the respective papers. We used our own implementation of the pFedBayes algorithm since the source code was not publicly available. As mentioned above, since the baselines assume the homogeneous models, we use larger VGG9 based models on the baselines.

\subsection{Results}
\begin{figure}[!t]
    \subfigure[]{\includegraphics[width=0.32\textwidth]{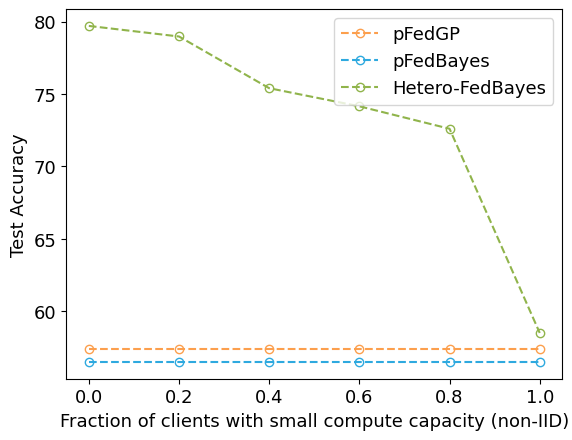}}
    \subfigure[]{\includegraphics[width=0.32\textwidth]{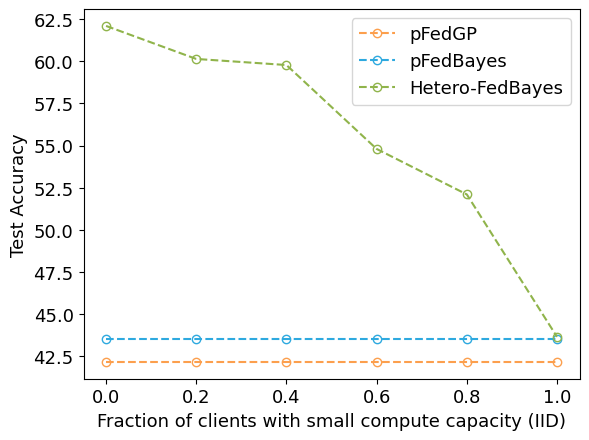}}
    \subfigure[]{\includegraphics[width=0.32\textwidth]{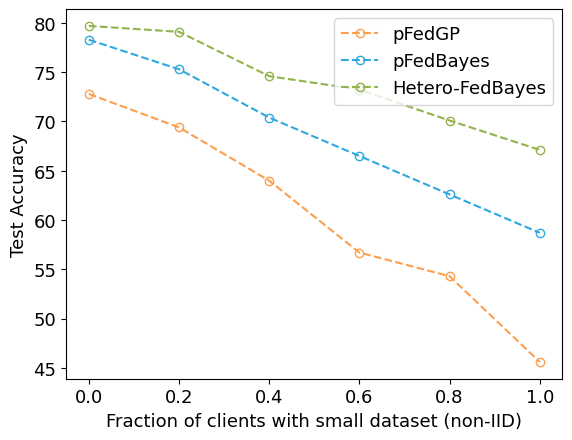}}
    \caption{Performance comparison of our method with baselines under different types of heterogeneity for CIFAR-10 dataset with 20 clients. Figure (a) is for heterogeneity in compute capacity across clients under non-IID data setting, figure (b) for compute heterogeneity under IID setting and figure (c) for heterogeneity in data resources. When a fraction of clients in the setting have low compute resource, the baselines being homogeneous can only train smaller models on all the clients as shown by constant performance. The results show that our method is more tolerant to both model heterogeneity and data heterogeneity across clients.}
    \label{fig:hetero}
\end{figure}

The performance of our method and the baselines under the non-IID data setting is reported in Table~\ref{table:acc_noniid}. Under the non-IID setting, we report the results corresponding to different dataset sizes on each client. To recall, in the small, medium and full settings each client has access to 50, 100 and 2500 training data points per class respectively. We observe that our method with homogeneous architectures across clients outperforms all other baselines. Moreover, when we consider the performance of our method under heterogeneous setting by considering 30\% of the total clients to be small capacity, it is evident that our method is better than the higher capacity homogeneous baselines for more complex tasks like in CIFAR-10 and CIFAR-100. On average, our method achieves about $6\%$ performance improvement over the baselines in the small and medium data settings. Figure~\ref{fig:hetero} compares the performance of our method with the highest performing baselines under various types of heterogeneity. Since our method can work with heterogeneous clients, we see that just by the proposed collaboration and having higher capacity clients in the FL ecosystem, the lower capacity clients are able to gain about $10\%$ increase in their performance. Also, the performance degradation of our method with change in the number of clients with limited data resources is more graceful as compared to the baselines. 

\section{Discussion} \label{sec:discussion}
We propose a novel method for Bayesian learning in heterogeneous federated learning settings and demonstrate that our method is able to outperform the baselines under different types of heterogeneous situations, while also providing a privacy guarantee and calibrated responses. The privacy analysis on the method provides us with various tunable knobs that can be adjusted to achieve the desired privacy-utility trade-off. The experiments show that the method is particularly useful for clients with lower data and lower compute resources as they can benefit the most by the presence of other, more powerful clients in the ecosystem. The availability of a small, unlabelled dataset at the server is typically a very mild requirement as such data can often be obtained from several open sources on the web. A direct future work will be to explore if suitable generative AI models could alternatively be used to readily provide application-specific alignment datasets. Secondly, as part of the future work the bound on the privacy loss random variable could be tightened by doing data specific characterization and modifying it for the specific local Bayesian training procedure which itself induces some amount of randomness.

\clearpage
\newpage
\small
\bibliography{refs}
\bibliographystyle{plainnat}

\newpage
\appendix
\begin{center}
{\textbf{ \Large Supplement for \enquote{Privacy Preserving Bayesian Federated Learning in Heterogeneous Settings}}}
\end{center}

In this supplementary material, we first include pseudo-code of the algorithm used for training our framework. Then, we discuss calibration metrics and present results demonstrating our method is well-calibrated. We also discuss the details about the alignment dataset, AD, its affect on the performance and the communication and computation cost of the procedure.

\section{Algorithm}
The pseudo-code of the algorithm used in the FedBNN method is included in the Algorithm~\ref{alg:algo}.

\begin{algorithm}[!htb]
   \caption{FedBNN Algorithm}
   \label{alg:algo}
   \begin{algorithmic}
   \STATE {\bfseries Input:} number of clients $N$, number of global communication rounds $T$, 
   number of local epochs $E$, weight vector $[w_1,w_2,\dots w_N]$, noise parameter $\gamma$, number of local training steps (before collaboration starts) $lsteps$
   \STATE {\bfseries Output:} Personalised BNNs $\{ \Phi_i | i \in [1, N] \}$ \\
   \STATE {\bfseries Server Side - }
   \FOR{$t=1$ {\bfseries to} $T$}
    \STATE AD = \textbf{X}
    \STATE Select a subset of clients $\mathcal{N}_t$  
    \FOR{each selected client $i \in \mathcal{N}_t$}
    \STATE $\Phi_i(\textbf{X}) =$ \textbf{LocalTraining}$( t, \Bar{\Phi}(\text{AD})^{(t-1)}, \textbf{X})$  
    \ENDFOR
   \STATE $\Bar{\Phi}(\text{AD})^{(t)} = \sum_{j=1}^{N} w_j \Phi_j(\textbf{X})$
   \ENDFOR
   \STATE Return {$\Phi_1(T), \Phi_2(T) \dots \Phi_N(T)$}
   
   \STATE {\bfseries LocalTraining}$(t, \Bar{\Phi}(\text{AD})^{(t-1)}, \textbf{X})$
   \IF{t < $lsteps$}
    \STATE{Continue}
   \ELSE
    \STATE{Generate $\Phi_{i}^{\text{corrected}} = \gamma \Bar{\Phi}(\textbf{X}) + (1 - \gamma) \Phi_{i}(\textbf{X})$}
    \FOR{each prior epoch}
     \STATE{Minimize CrossEntropy($\Phi_{i}^{\text{corrected}}, \Bar{\Phi}(\text{AD})^{(t-1)})$ to obtain prior parameters $\psi$ of the BNN $\Phi_{i}$}
    \ENDFOR
   \ENDIF   
   \FOR{each local epoch}
    \STATE{Minimize $\text{KL}[q(\W_i | \theta) || p(\W_i; \psi^*)] - \E_{q(\mathcal{W}_i | \theta)}[log \probP(\Xx_i | \W_i)]$} over $\{\theta : q(\mathcal{W}_i | \theta) \in \mathcal{Q}\}$ to obtain $\theta^*$
    \ENDFOR
   \STATE{Obtain $\Phi_i(\textbf{X})$ by Monte-carlo sampling from $\probP(\W_i |\Xx)$}
   \STATE Return $\Phi_i(\textbf{X})$

\end{algorithmic}
\end{algorithm}

\section{Calibration}
\begin{figure}[!htb]
    \subfigure[\textbf{Dataset}: CIFAR-10, \textbf{ECE}: 0.070, \textbf{MCE}: 0.134]{\includegraphics[width=0.45\textwidth]{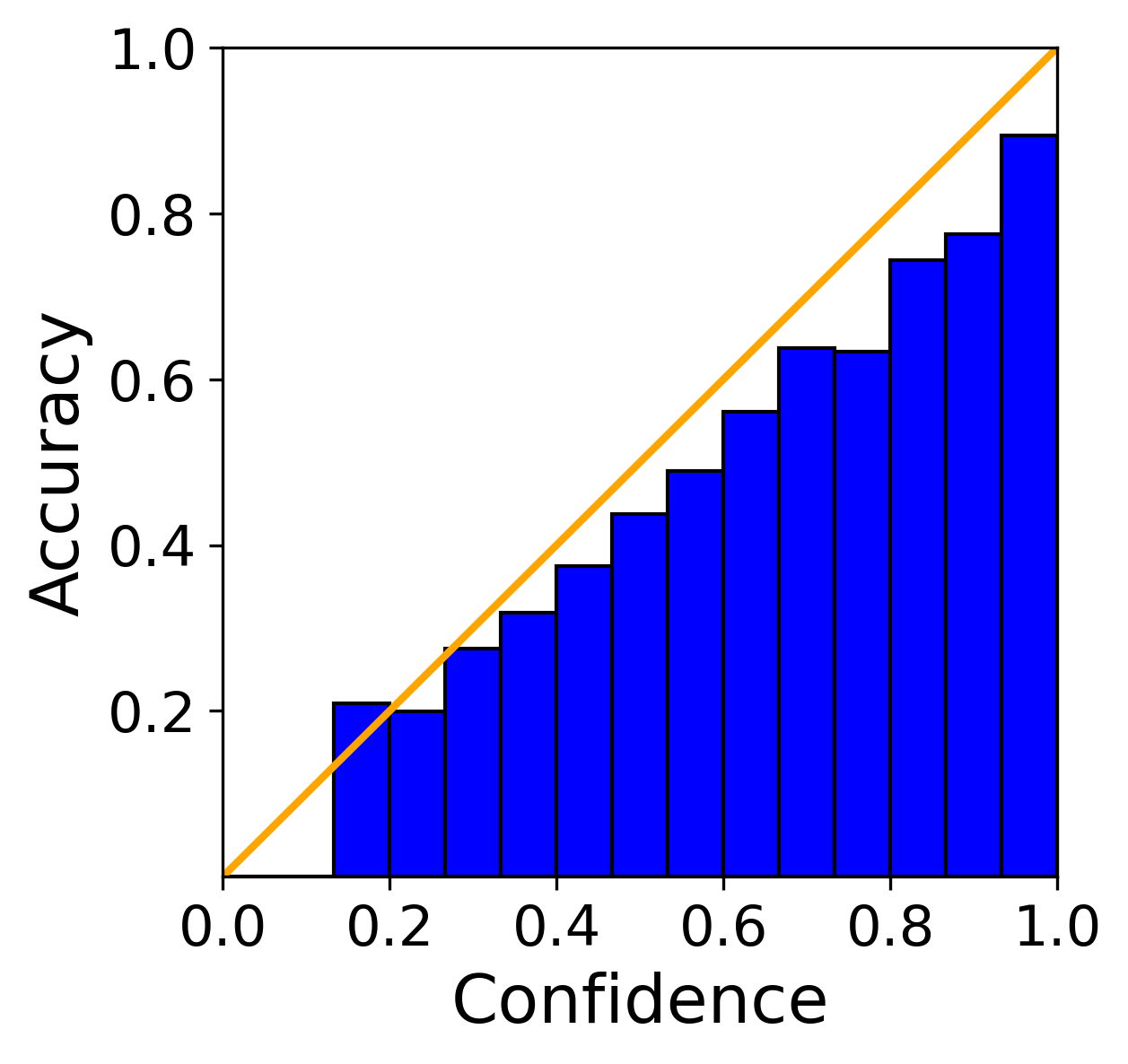}}
    \subfigure[\textbf{Dataset}: MNIST, \textbf{ECE}: 0.032, \textbf{MCE}: 0.156]{\includegraphics[width=0.45\textwidth]{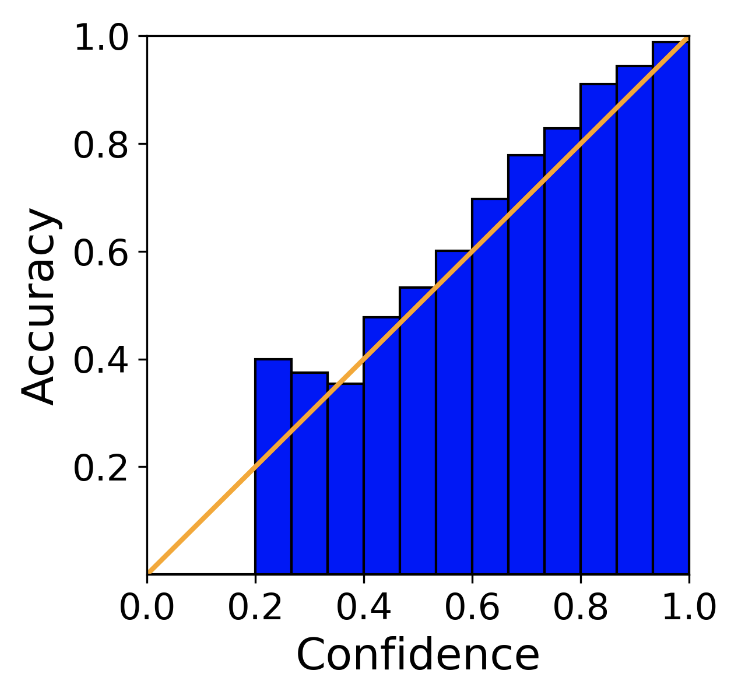}}
    \caption{Reliability diagrams and scores showing model calibration. Figure (a) is for the results corresponding to the CIFAR-10 dataset and Figure (b) for MNIST dataset.}
    \label{fig:calibration}
\end{figure}

Model calibration is a way to determine how well the model's predicted probability estimates the model's true likelihood for that prediction. Well-calibrated models are much more important when the model decision is used in critical applications like health, legal etc. because in those cases managing risks and taking calculated actions require a confidence guarantee as well. Visual tools such as reliability diagrams are often used to determine if a model is calibrated or not. In a reliability diagram, model's accuracy on the samples is plotted against the confidence. A perfectly calibrated model results in an identity relationship. Other numerical metrics that could be used to measure model calibration include Expected Calibration Error (ECE) and Maximum Calibration Error (MCE). ECE measures the expected difference between model confidence and model accuracy whereas MCE measures the maximum deviation between the accuracy and the confidence. The definitions and empirical formulas used for calculating ECE and MCE are as given below.
$$
\text{ECE} = \E_{\hat{P}}[\probP(\hat{Y} = Y | \hat{P} = p) - p]
$$
$$
\text{MCE} = \max_{p \in [0,1]} | \probP(\hat{Y} = Y | \hat{P} = p) - p |
$$
Emprically,
$$
\text{ECE} = \sum_{i=1}^{M} \frac{|B_i|}{n} |\text{accuracy}(B_i) - \text{confidence}(B_i)|
$$
$$
\text{MCE} = \max_{i \in [1,M]} |\text{accuracy}(B_i) - \text{confidence}(B_i)|
$$
where $B_i$ is a bin with set of indices whose prediction confidence according to the model falls into the range $\big( \frac{i-1}{M} , \frac{i}{M} \big)$. Figure~\ref{fig:calibration} shows the reliability diagram along with the ECE and MCE scores for our method measured on MNIST and CIFAR-10 dataset in the non-IID data setting.

\section{Alignment Dataset (AD)}
 
\begin{figure}[!htb]
    \begin{center}
    \includegraphics[scale=0.3]{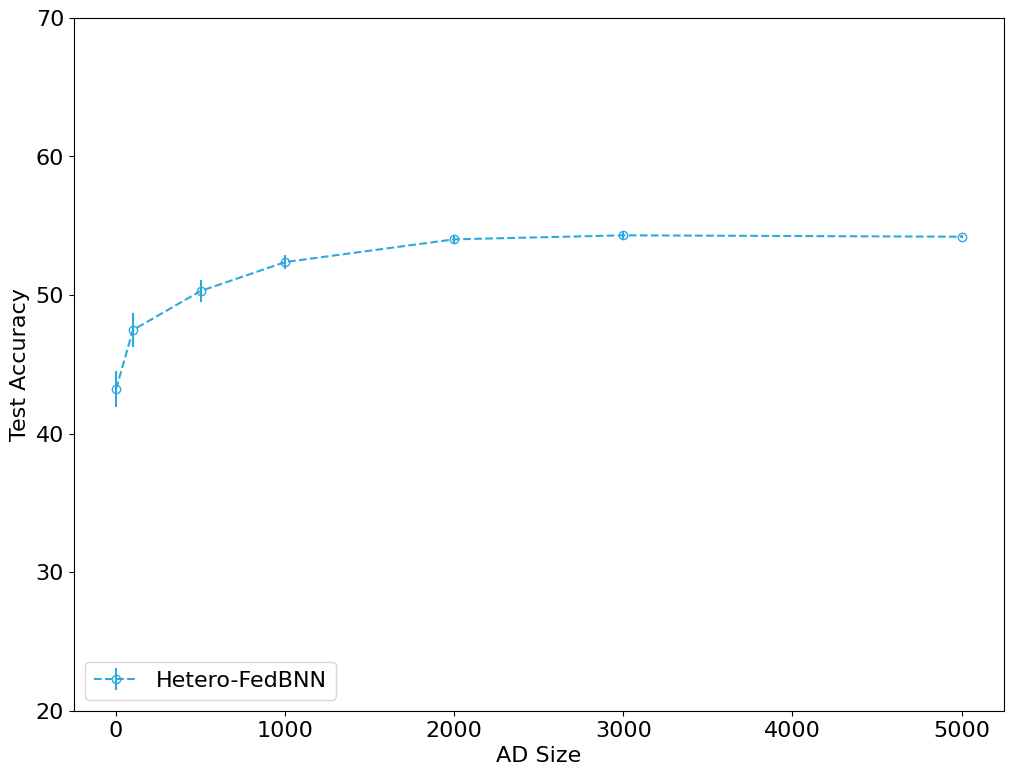}
    \end{center}
    \caption{Ablation study comparing effect of AD size on the performance. The included results are for CIFAR-10 dataset in the small data setting with non-IID partitions and heterogeneous clients.}
    \label{fig:rad_size2}
\end{figure}

In FedBNN, the alignment dataset is used to achieve collaboration across clients. Since the AD is assumed to be of the same domain as the target application, there is no practical constraint on obtaining the AD in real-world settings. In many cases it could be obtained from web, for example images from common datasets in Huggingface, texts from Wikipedia, Reddit etc. The use of AD is not different from how several other methods use an additional dataset for augmentation. The effect of size of AD on the performance of models is demonstrated in Figure~\ref{fig:rad_size2} for CIFAR-10 dataset in the small data and non-IID setting. In that figure, we observe that when the size of AD is small the performance of the model is low but as the size of AD increases the performance increases up to a point and becomes constant afterwards. The number of data points in AD that are required to achieve good improvement in the model performance is small and practical.

\section{Communication and Computation Efficiency}
\paragraph{Communication Cost} In FedBNN, each global communication round requires that the server sends the alignment dataset to all the clients and the clients upload the outputs of their respective models on the common dataset AD. Since AD is a publicly available dataset, AD could be transmitted to the clients by specifying the source and the indices, and does not really needs to be communicated across the channel. The client output on AD, on the other hand, depends on the number of instances in AD, let's call it $K$, therefore, the total communication cost in each round of our method is $O(K)$. As shown in Figure~\ref{fig:rad_size2}, having $K = 2000$ gives a good performance. The communication cost between the clients and the server, thus, is also invariant of the number of model parameters which tend to run in millions. This allows our method to be much more communication efficient as compared to the conventional FL algorithms that transmit model parameters in each communication round, making it practically more useful.

\paragraph{Computation Cost} Similarly, the computation cost of a FL procedure involves the costs incurred in local training at the individual clients and the cost of aggregation at the server. When clients are training Bayesian models, the local training cost depends on the inference method used for obtaining the posterior distributions over all model parameters. We choose variational inference (VI) based procedure to obtain the  weight distributions since VI based methods are computationally more efficient. Also, since we do not aggregate the posterior distributions over parameters at the server, the aggregation cost does not depend on the number of model parameters making it an efficient procedure.


\end{document}